\documentclass[sigconf, screen, nonacm]{acmart}
\renewcommand\footnotetextcopyrightpermission[1]{} 
\AtBeginDocument{%
  }

\setcopyright{acmlicensed}
\copyrightyear{2018}
\acmYear{2018}
\acmDOI{XXXXXXX.XXXXXXX}
\acmConference[Conference acronym 'XX]{Make sure to enter the correct
  conference title from your rights confirmation email}{June 03--05,
  2018}{Woodstock, NY}
\acmISBN{978-1-4503-XXXX-X/2018/06}

\usepackage{amsmath}
\usepackage{mathtools}
\usepackage{amsthm}
\usepackage{multirow}
\usepackage{makecell}
\usepackage{nicefrac}
\usepackage{colortbl}   
\usepackage{xcolor}
\usepackage{wrapfig}
\usepackage{tcolorbox}
\tcbuselibrary{breakable}
\begin{document}

\title{Perceive, Verify and Understand Long Video: Multi-Granular Perception and Active Verification via Interactive Agents}

\author{
Jiahua Li$^{1}$, Zhanhe Zhang$^{1}$, Chenghao Xu$^{3}$, Zhe Xu$^{2}$, Kun Wei$^{1}$, Xu Yang$^{1}$, Cheng Deng$^{1}$ \\
$^{1}$School of Electronic Engineering, Xidian University, Xi'an, China \\
$^{2}$Department of Computer Science and Engineering, Hong Kong University of Science and Technology, Hong Kong, China \\
$^{3}$College of Computer and Information, Hohai University, Nanjing, China \\
\texttt{\{ljhxdu, weikunsk, zhexu.xd, xuyang.xd, chdeng.xd\}@gmail.com} \\
}

\renewcommand{\shortauthors}{Li et al.}

\begin{abstract}
Long videos, characterized by temporal complexity and sparse task-relevant information, pose significant reasoning challenges for AI systems. Although existing Large Language Model (LLM)-based approaches have advanced long video understanding, they remain bottlenecked by task-agnostic, fixed-granularity perception pipelines and suffer from vision-language hallucinations. Inspired by human adaptive perception and active verification, we propose CogniGPT, a framework leveraging an interactive loop between a Multi-Granular Perception Agent (MPA) and an Active Verification Agent (AVA). Specifically, instead of predetermined heuristics, MPA adaptively determines the optimal perception granularity and strategy based on the evolving context, while AVA actively mines multi-perspective visual evidence to cross-verify key observations and eliminate hallucinations. This interaction allows CogniGPT to efficiently identify a minimal set of reliable task-related clues. Extensive experiments on EgoSchema, Video-MME, NExT-QA, and MovieChat demonstrate its superiority in accuracy and efficiency. Notably, on EgoSchema, it surpasses existing training-free methods using only 11.2 frames and achieves performance comparable to Gemini 1.5-Pro.
\end{abstract}




\maketitle

\section{Introduction}

Benefiting from the intricate mechanisms of brain and cognition, human intelligence fundamentally excels at effortlessly comprehending complex multimodal content (including natural language and hours-long videos) while performing sophisticated reasoning. However, this capability poses significant challenges for artificial intelligence systems, particularly in understanding and reasoning about dynamic visual content alongside textual information. With advancements in computer vision, various multimodal video tasks have been extensively studied, such as video question answering~\citep{bai2023glance,kim2023semi}, moment retrieval~\citep{moon2023query,xu2024exploiting}, and video captioning~\citep{zhao2023learning}. Nevertheless, these tasks typically focus on isolated aspects of video analysis and struggle to achieve comprehensive multimodal understanding and reasoning, remaining far from realizing Artificial General Intelligence.

\begin{figure}[t]
\centering
\includegraphics[width=0.99\linewidth]{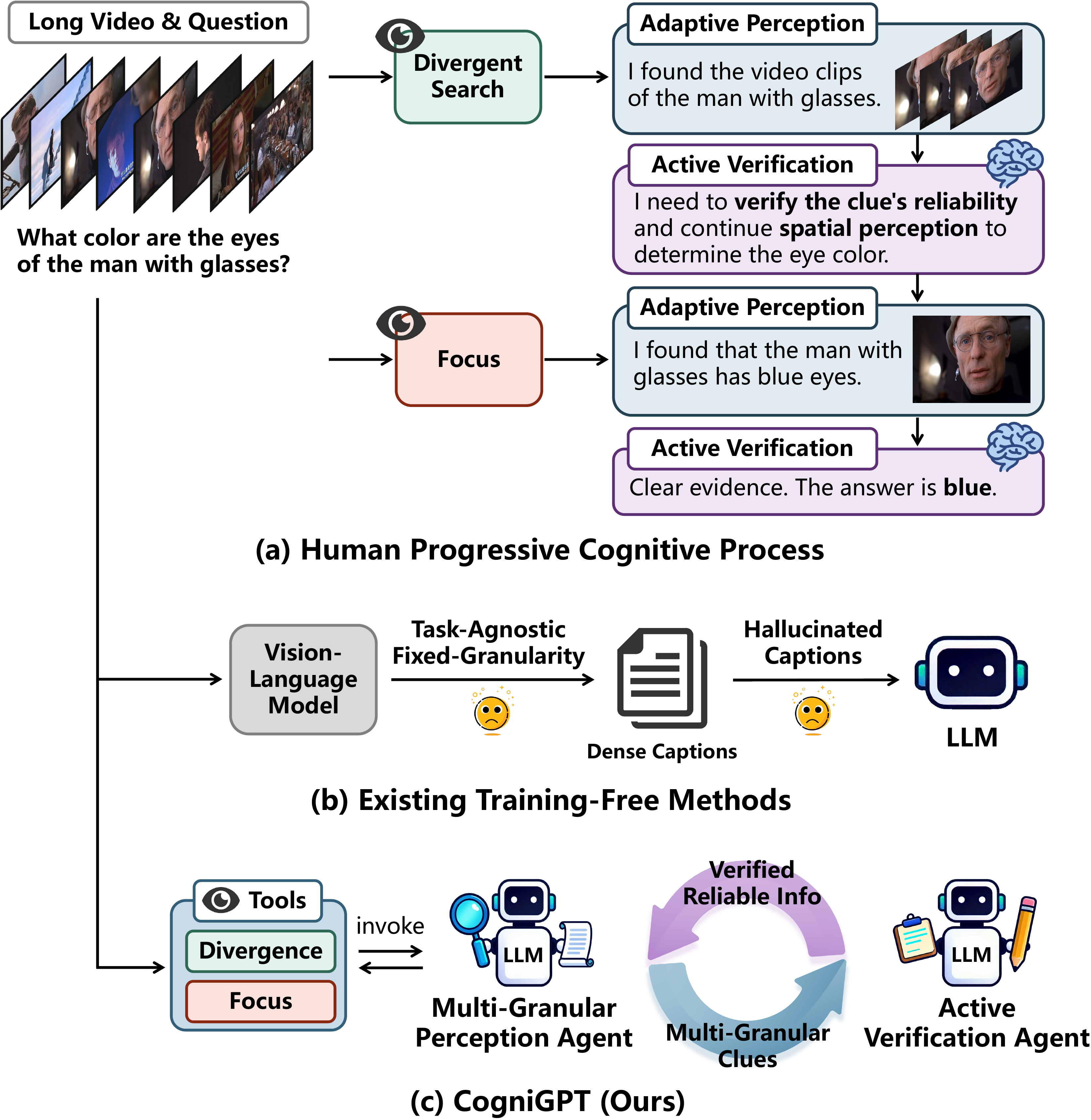}
\caption{(a) Humans comprehend long videos through a progressive process of adaptive perception and active verification.
(b) Existing training-free methods are bottlenecked by task-agnostic, fixed-granularity perception pipelines and suffer from vision-language hallucinations.
(c) CogniGPT (Ours) employs an interactive loop between MPA and AVA, combining adaptive multi-granular perception with active verification to efficiently explore reliable clues.}
\label{fig_intro}
\end{figure}

With the rapid advancement of Large Language Models (LLMs), harnessing their commonsense knowledge and reasoning capabilities for complex video understanding has garnered increasing attention. Existing approaches can be broadly categorized into two paradigms: Multimodal Large Language Model (MLLM)-based methods and LLM Agent-based methods. MLLM-based methods~\citep{li2023videochat,zhang2023video,maaz2023video} align visual and textual tokens through end-to-end training, enabling LLMs to comprehend video content. However, these methods encounter substantial challenges when processing long videos, primarily due to the difficulty of effectively balancing spatial-temporal details and capturing long-range dependencies within a limited number of visual tokens. 
Although some MLLM-based methods~\citep{song2024moviechat,he2024ma} adopt various compression strategies to partially alleviate these issues, they still suffer from hallucinations~\citep{ma2024vista}, incur high training costs, and lack interpretable reasoning capabilities for complex tasks.

More recently, several pioneering works attempt to use LLMs as agents for video understanding~\citep{fan2024videoagent, zhang2023simple, ma2024drvideo}, demonstrating significantly enhanced reasoning capabilities in a training-free manner. However, they remain bottlenecked by \textbf{fixed, single-granularity perception pipelines}. Specifically, whether relying on dense captions (e.g., LLoVi~\citep{zhang2023simple} and DrVideo~\citep{ma2024drvideo}), fixed clustering strategies (e.g., VideoTree~\citep{wang2025videotree}), or predefined retrieval mechanisms (e.g., VideoAgent*~\citep{wang2024videoagent}), these methods cannot adaptively adjust their perception granularity and strategy based on task complexity. Consequently, they are inefficient and prone to overlooking sparse, task-relevant information. Furthermore, their reasoning is easily misled by \textbf{vision-language hallucinations within the captions}, as they lack effective mechanisms to eliminate such hallucinations and prevent them from interfering with the reasoning process.

To address the aforementioned limitations, we turn to how humans perceive and understand videos. Unlike rigid perception pipelines, humans do not apply a fixed observation pattern. Instead, they adaptively determine their perception granularity and strategy based on task complexity and current information. Furthermore, rather than passively trusting initial visual impressions, humans actively verify critical information to eliminate ambiguities and prevent misinterpretation. Through an iterative loop of adaptive perception and active verification, humans progressively explore a minimal set of reliable clues, enabling highly efficient and accurate reasoning.

Inspired by this insight, we propose CogniGPT, a novel framework that establishes an interactive loop of adaptive perception and active verification. Specifically, to break away from fixed, single-granularity pipelines, we design the Multi-Granular Perception Agent (MPA). Leveraging the powerful planning capabilities of LLMs, \textbf{MPA adaptively determines the optimal perception granularity and strategy based on task complexity and current information}---ranging from broad divergent exploration to fine-grained spatiotemporal focus---to efficiently extract task-relevant clues. Furthermore, to prevent vision-language hallucinations from interfering with the reasoning process, we introduce the Active Verification Agent (AVA). \textbf{AVA actively explores complementary information from different perspectives and granularities to cross-verify key observations}, eliminating hallucinations while providing verbal feedback to guide MPA's next steps. Ultimately, through the synergy of these two agents, CogniGPT achieves highly accurate long-video understanding by exploring a minimal set of frames.
Extensive experiments on the EgoSchema~\citep{mangalam2023egoschema}, Video-MME~\citep{fu2024video}, MovieChat~\citep{song2024moviechat}, and NExT-QA~\citep{xiao2021next} benchmarks demonstrate the superiority of CogniGPT in both accuracy and efficiency. Notably, on the EgoSchema benchmark, CogniGPT surpasses the accuracy of DrVideo~\citep{ma2024drvideo} by 2.8\%, while using only 12.4\% of the frames and 27.6\% of the runtime. Furthermore, it performs competitively with the proprietary Gemini 1.5-Pro.
In summary, the key contributions of this work are as follows:

\begin{itemize}
\item We propose \textbf{CogniGPT}, an interactive perception-verification framework that progressively explores minimal yet reliable clues for long video understanding.

\item We design \textbf{MPA} to break through the constraints of fixed pipelines by adaptively adjusting perception granularities and strategies based on task complexity, while introducing \textbf{AVA} to cross-verify key observations across multiple perspectives and granularities to eliminate hallucinations.

\item Extensive experiments on challenging benchmarks, including Video-MME and EgoSchema, demonstrate that CogniGPT achieves state-of-the-art performance in both accuracy and efficiency.
\end{itemize}

\section{Related Work}

\subsection{MLLMs for Long Video Understanding}

Long video understanding is challenging due to spatio-temporal complexity and the sparsity of relevant information. The key challenge for MLLM-based methods lies in balancing spatial-temporal details and capturing long-range dependencies within a limited context window. As a result, these methods focus on token compression strategies~\citep{hussein2019videograph,islam2022long,nguyen2022s4nd,song2024moviechat,wang2024videoagent}. For example, MovieChat~\citep{song2024moviechat} reduces token redundancy, and Chat-UniVi~\citep{jin2024chat} applies kNN clustering for compression. 
Despite partially mitigating these issues, MLLM-based methods still suffer from hallucinations~\citep{ma2024vista}, high training costs, and a lack of interpretable reasoning. In contrast, we employ LLMs as agents to emulate a human-like \textit{Perception-Verification} loop, progressively exploring reliable clues for long video understanding in a training-free manner.



\subsection{LLM Agents for Long Video Understanding}

Recent training-free agents have significantly advanced long video understanding, yet they remain limited by \textbf{fixed, single-gran\-u\-lar\-i\-ty perception pipelines}. Specifically, LLoVi \citep{zhang2023simple} and DrVideo \citep{ma2024drvideo} rely on global dense captions, VideoAgent*~\citep{wang2024videoagent} employs predefined retrieval, and VideoTree~\citep{wang2025videotree} utilizes fixed clustering strategies. Such rigid strategies cannot adaptively adjust perception based on task difficulty, resulting in either redundant processing of irrelevant frames or the omission of critical clues limited by the quality of retrieval, clustering, and captioning. Furthermore, these methods are highly susceptible to \textbf{vision-language hallucinations}. Since they lack an active mechanism to verify and eliminate false information in the passively generated text, their reasoning is easily misled.

To address these limitations, we propose CogniGPT, an interactive loop of adaptive perception and active verification. Within this framework, MPA transcends fixed, single-granularity pipelines by adaptively determining optimal perception strategies and granularities based on task complexity and current information. To ensure reasoning reliability, AVA cross-verifies key observations across multiple perspectives and granularities to eliminate hallucinations and prevent them from misleading the reasoning process.

\begin{figure*}[t]
    \centering
    \includegraphics[width=0.95\textwidth]{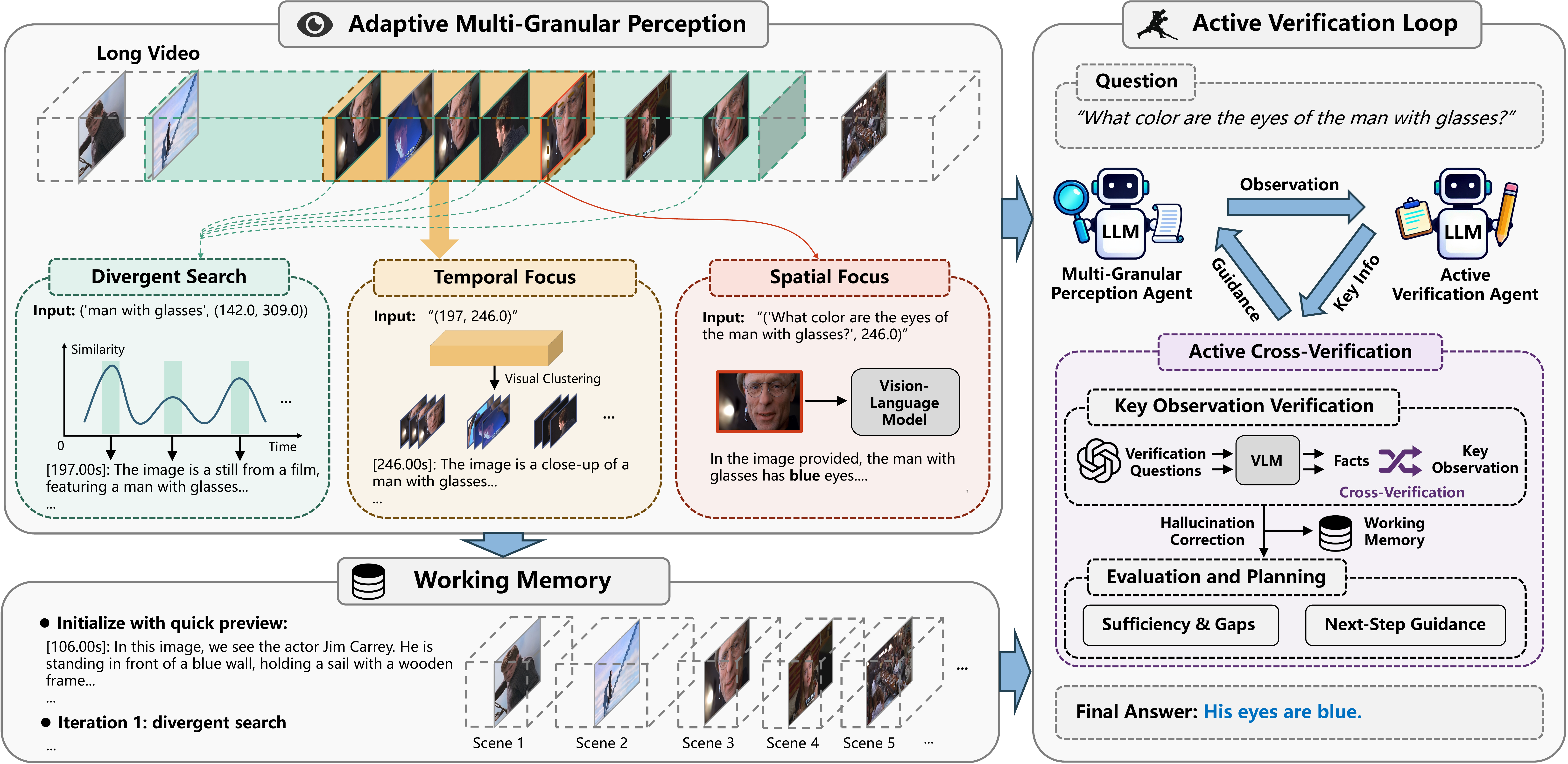}
    \caption{
    Overview of CogniGPT. \textbf{Left:} \textit{Adaptive Multi-Granular Perception} adaptively extracts key clues from divergent and focused perspectives based on current information. \textbf{Right:} The \textit{Active Verification Loop} drives progressive exploration via an interactive loop between the Multi-Granular Perception Agent (MPA) and the Active Verification Agent (AVA), eliminating hallucinations through multi-perspective active cross-verification.
    }
    \label{overview}
\end{figure*}

\section{Methods}

\subsection{Overview}

Given a long video $V$ and a complex query $Q$, we propose CogniGPT to efficiently explore a minimal set of reliable clues through an interactive loop between a Multi-Granular Perception Agent (MPA) and an Active Verification Agent (AVA), as shown in Figure~\ref{overview}. Specifically, at each iteration $t$, MPA $\pi_p$ adaptively determines the perception granularity based on the current task and evolving information. It achieves this by selecting an action $a_t$ and its corresponding input, taking into account the query $Q$, the current \textit{Working Memory} $\mathcal{M}_{t-1}$, and the guidance $g_{t}$ provided by AVA, following the policy $\pi_p(a_t \mid Q, \mathcal{M}_{t-1}, g_{t})$. The selected action produces an observation $o_t$, which is incorporated into memory to update the state $\mathcal{M}_t$. To ensure reliable reasoning, AVA then performs active cross-verification to validate critical clues and strictly eliminate hallucinations, while providing verbal feedback to refine the subsequent perception strategy. We detail each component below.

\subsection{Adaptive Multi-Granular Perception}

To break the bottleneck of fixed-granularity pipelines, MPA is designed to adaptively determine the optimal perception granularity based on task complexity and current information. To achieve this, we construct a dynamic action space that empowers the LLM to autonomously orchestrate its perception strategy across multiple granularities. The defined actions include \texttt{divergent search}, \texttt{temporal focus}, and \texttt{spatial focus}, described as follows:

\textbf{Divergent Search.}
We simulate the divergent search mechanism in human visual attention through the \texttt{divergent search} action, designed to identify key frames relevant to a subtask over a broad temporal range, avoiding redundant video captioning.
Specifically, the LLM first infers a subtask query $q$ along with a target temporal span as the input to this action. For example, if the main query $Q$ is ``What color are the eyes of the man with glasses?'', the LLM may extract ``man with glasses'' as the sub-query $q$.

We employ EVA-CLIP-8B~\citep{sun2024eva} to extract the textual representation of $q$ and the visual representations of sampled video frames $v_i$ within the specified span. We then compute cosine similarity scores $s_i$ between $q$ and each $v_i$.
To emulate the human divergent search, we select frames with broad contextual diversity, inspired by the watershed algorithm~\citep{vincent1991watersheds}. Specifically, we smooth the similarity scores ${s_1, s_2, \ldots, s_T}$ using a sliding-window average to obtain refined scores $\tilde{s}_i = \text{smooth}(s_i)$. We then compute the mean similarity $\bar{s} = \frac{1}{T} \sum_{i=1}^{T} \tilde{s}_i$ as a threshold to segment the timeline into peak and valley regions. From each peak region, we select the frame with the highest $\tilde{s}_i$ score and retain the top-$N_f$ frames overall. Unlike existing methods~\citep{fan2024videoagent} that use a top-k selection strategy and often pick redundant frames clustered around the highest peaks, our approach captures diverse, causally relevant contextual information, reducing retrieval redundancy.

After retrieving the $N_f$ key frames, we use a vision-language model to generate captions for each selected frame, along with their timestamps. This action enables the LLM to efficiently locate and understand query-relevant visual evidence, including temporal aspects such as ``when'' certain events occur.

\textbf{Temporal Focus.}
While \texttt{divergent search} effectively retrieves relevant video segments, its coarse temporal granularity can overlook crucial fine-grained details like rapid action transitions. To address this, we introduce \texttt{temporal focus} as a complement, establishing a coarse-to-fine reasoning paradigm. After \texttt{divergent search} identifies a broad span, \texttt{temporal focus} performs a granular semantic analysis within it, capturing key sub-events and enabling a detailed understanding of dynamic content.

Specifically, to balance granularity and efficiency, we first employ EVA-CLIP-8B~\citep{sun2024eva} to extract video features within the given temporal span. We then perform K-means clustering based on semantic similarity, obtaining $K_t$ clusters ${C_1, \dots, C_{K_t}}$. For each cluster $C_i$, we select the clip nearest to its centroid $c_i$ as the representative and generate a caption using a vision–language model. This process yields $K_t$ captions that summarize the representative semantic content of the segment, thereby enabling fine-grained yet efficient temporal understanding.

\textbf{Spatial Focus.}
Although we utilize \texttt{divergent search} and \texttt{temporal focus} to obtain the captions for frames or clips, the understanding of spatial content heavily relies on the quality of the captions. This reliance may result in the inability to capture fine-grained spatial details, attributes, and positional relationships. For instance, as shown in Figure~\ref{overview}, when the question is ``What color are the eyes of the man with glasses?'', the caption for the relevant frame does not include information about the color of the eyes, necessitating further spatial understanding. To address this issue, we propose the \texttt{spatial focus} action, which enables fine-grained spatial understanding through Visual Question Answering (VQA). Specifically, we prompt the LLM to input the frame to be analyzed and a sub-task question, and leverage LLaVA-NeXT~\citep{liu2024llavanext} to perform VQA on the frame, enabling the extraction of task-relevant spatial information that goes beyond what is captured in the initial caption.

It is worth emphasizing that although we perform global video feature extraction with EVA-CLIP-8B, along with similarity computation and clustering, the required runtime is negligible compared to caption generation and LLM inference. Supporting experimental evidence is presented in Section~\ref{sec: Runtime Analysis}.
The detailed prompt descriptions of the actions are provided in the Appendix.

\subsection{Working Memory}

Inspired by human working memory, we propose a dynamically updated \textit{Working Memory} $\mathcal{M}$ to store perceived information, thus providing an evolving context for LLM-based planning and reasoning.

\textbf{Initialization with Quick Preview.} Before engaging in reasoning, humans typically form a general understanding of the video context. In a similar manner, we employ EVA-CLIP-8B~\citep{sun2024eva} to extract video features, perform K-means clustering into $K_m$ clusters, and generate captions for the corresponding cluster centroids.
These $K_m$ captions summarize the representative scenes within the video. The selected frames' timestamps and captions are integrated into $o_0$ to initialize $\mathcal{M}$, supplying essential contextual cues for LLM reasoning.

\textbf{Dynamic Update.} At each iteration $t$, the new action $a_t$ and observation $o_t$ are added to memory $\mathcal{M}$, forming $\mathcal{M}_t = (a_0, o_0, a_1, o_1,\allowbreak \dots, a_t, o_t)$, which provides the LLM with progressively richer contextual cues.

\subsection{Active Verification}

Given that MPA is susceptible to hallucinations from vision-language models (VLMs) and exhibits limited planning capabilities, often leading to suboptimal strategies, we propose the Active Verification Agent (AVA) to continuously detect and correct hallucinated perceptual information. Furthermore, it analyzes the global context and provides verbal feedback to optimize the action strategy of MPA.

\textbf{Active Cross-Verification.}
Since reasoning depends on textual descriptions generated by the VLM, hallucinations in key observations may directly cause incorrect inference by the LLM. To address this issue, inspired by Chain-of-Verification~\citep{dhuliawala2023chain}, AVA performs cross-verification of critical observations. Specifically, at each iteration step $t$, AVA first determines whether the latest observation $o_{t-1}$ contains key information relevant to answering the question $Q$. 
If such information exists, the verification proceeds as follows: we prompt the LLM to freely generate a set of verification questions (2-3 in our experiments) from multiple perspectives based on the identified key information across frames. For example, if the original question $Q$ is ``What color is the boy's hat in the video?'' and a caption in a certain frame states ``a boy is wearing a red hat,'' then a possible verification question is ``What color is the boy's hat in the image?'' These verification questions together with their corresponding frames are input into the VLM for visual question answering, yielding multiple factual responses. The LLM is then prompted to perform cross-verification between these factual responses and the original key information to assess its reliability. If the information is deemed trustworthy, it is stored in the \textit{Working Memory}. This verification mechanism essentially leverages complementary multi-frame information—capturing different viewpoints and granularities—to eliminate hallucinatory content.

\textbf{Evaluation and Planning.}
Subsequently, AVA performs a chain-of-thought (CoT) analysis over the question $Q$ and the current \textit{Working Memory} $\mathcal{M}_{t-1}$, evaluating key aspects, including: \textit{Information Sufficiency}, which examines whether the available information is sufficient to reliably answer the question; \textit{Information Gaps}, identifying any critical missing details; and \textit{Next Step Decision}, determining whether the agent should proceed with further information gathering or conclude with a final answer.
If the decision is to continue, AVA generates a guidance output $g_{t}$, which specifies the next piece of information to collect and suggests the most effective actions for doing so. This guidance is then provided as verbal feedback to MPA. 
If the decision is to terminate, AVA offers a final explanation and provides the answer. Additionally, a maximum iteration limit $T_{\text{max}}$ is set, upon which a final answer is produced based on the currently available information.

\begin{table*}[t]
    \centering
    \caption{Results on the EgoSchema, Video-MME long split, NExT-QA, and MovieChat benchmarks. Accuracy is reported in \%, and ``Frame'' denotes the average number of frames per sample that require captioning and QA.}
    \label{tab:main1}
    \begin{tabular}{lccccccccc}
      \toprule
      \multirow{2}{*}{Method} & \multirow{2}{*}{LLM} &
      \multicolumn{2}{c}{EgoSchema} &
      \multicolumn{2}{c}{Video-MME} &
      \multicolumn{2}{c}{NExT-QA} &
      \multicolumn{2}{c}{MovieChat} \\
      \cmidrule(lr){3-4} \cmidrule(lr){5-6} \cmidrule(lr){7-8} \cmidrule(lr){9-10}
       &  & Acc. & Frames & Acc. & Frames & Acc. & Frames & Acc. & Frames \\
      \midrule
      \multicolumn{10}{l}{\textbf{Training-Free LLM Agent}} \\
      LLoVi~\citep{zhang2023simple} & GPT-3.5 & 51.8 & 180 & 45.4 & 492 & 67.7 & 22 & 58.3 & 180 \\
      VideoAgent*~\citep{wang2024videoagent} & GPT-4 & 60.2 & \textbf{8.4} & 46.4 & 24.6 & 71.3 & \textbf{8.2} & 58.1 & -- \\
      VideoAgent~\citep{fan2024videoagent} & GPT-4 & 62.8 & $>$90 & -- & -- & 70.8 & 22 & -- & -- \\
      VideoTree~\citep{wang2025videotree} & GPT-4 & 66.2 & 62.4 & \underline{53.1} & 128.0 & 73.5 & 12.6 & -- & -- \\
      DrVideo~\citep{ma2024drvideo} & GPT-4 & 66.4 & $>$90 & 51.7 & $>$492 & -- & -- & 93.1 & $>$492 \\
      \rowcolor{gray!20}
      Ours & GPT-4 & \textbf{69.2} & \underline{11.2} & \textbf{54.7} & \textbf{18.3} & \textbf{76.8} & \underline{10.9} & \textbf{95.3} & \textbf{10.5} \\
      \bottomrule
    \end{tabular}
\end{table*}

\begin{table}[t]
  \centering
  \small
  \caption{Comparison with Proprietary and Open-Source MLLMs on EgoSchema and Video-MME long split. IV2.5 and Q2.5 denote the underlying VLM backbones InternVL2.5 and Qwen2.5-VL, respectively.}
  \label{tab:EgoSchema_all}

  \resizebox{\linewidth}{!}{
  \begin{tabular}{l cc cc}
    \toprule
    \multirow{2}{*}{Method} & \multicolumn{2}{c}{EgoSchema} & \multicolumn{2}{c}{Video-MME} \\
    \cmidrule(lr){2-3} \cmidrule(lr){4-5}
     & Acc. & Frames & Acc. & Frames \\
    \midrule
    \textbf{Proprietary} \\
    Gemini 1.5-Flash & 65.7 & 180 & 61.1 & 1230 \\
    Gemini 1.5-Pro   & 71.2 & 180 & \textbf{67.4} & 1230 \\
    GPT-4o           & \textbf{72.2} & 180 & 65.3 & \textbf{384} \\
    \midrule
    \textbf{Open-Source} \\
    LLaVA-OneVision-72B & 62.0 & \textbf{32} & 60.0 & 32 \\
    Qwen2.5-VL-7B       & 65.0 & 180 & -- & -- \\
    Qwen2.5-VL-72B      & \textbf{76.2} & 180 & -- & -- \\
    VideoChat-A1 (IV2.5)& 72.1 & 35.9 & 63.7 & 35.9 \\
    VideoChat-A1 (Q2.5) & 70.7 & 42 & 64.3 & 42 \\
    \midrule
    \textbf{Training-Free Agent} \\
    \rowcolor{gray!15}
    Ours (IV2.5-8B) & \textbf{74.0} & 11.5 & 66.3 & \textbf{20.4} \\
    \rowcolor{gray!15}
    Ours (Q2.5-7B)  & 72.4 & \textbf{10.7} & \textbf{67.0} & 23.8 \\
    \bottomrule
  \end{tabular}
  }
\end{table}

\section{Experiments}

\subsection{Datasets and Metrics}

We evaluate our approach on four benchmarks: EgoSchema \citep{mangalam2023egoschema}, with 500 samples from the official subset averaging 3 minutes; Video-MME~\citep{fu2024video}, using the long split of the 300-video open-domain subset averaging 41 minutes; NExT-QA~\citep{xiao2021next}, with 600 instances across causal, temporal, and descriptive types averaging 44 seconds; and MovieChat~\citep{song2024moviechat}, using the global mode of the official test split with 170 samples averaging 10 minutes. These datasets span diverse durations, question types, and reasoning tasks, enabling a comprehensive evaluation of CogniGPT. Full details are provided in the Appendix.

For EgoSchema, NExT-QA, and Video-MME, we evaluate accuracy on multiple-choice questions. 
For MovieChat, we use GPT-assisted accuracy evaluation (true/false). Following DrVideo~\citep{ma2024drvideo}, we select Gemini-Pro~\citep{team2023gemini} as the evaluation assistant and adopt the same prompt~\citep{maaz2023video} for a fair comparison.
Considering that captioning and QA are the primary sources of runtime, to compare the efficiency of training-free LLM Agent approaches, we also report the average number of frames requiring captioning and QA per sample.

\subsection{Implementation Details}

To balance efficiency and accuracy, we configure CogniGPT with the following settings: 
On EgoSchema, NextQA, and MovieChat, we set $N_f$ = 5, $K_t$ = 3, and $K_m$ = 5; on VideoMME, we set $N_f$ = 8, $K_t$ = 5, and $K_m$ = 5. The maximum number of interaction iterations is set to $T_{\max}$ = 3. We preprocess the raw videos at 1 FPS for the EgoSchema and NExT-QA benchmarks, 0.5 FPS for MovieChat, and 0.125 FPS for Video-MME. All experiments are conducted on 4 (or fewer) NVIDIA A6000 GPUs. Error analysis is provided in the Appendix. 

\subsection{Main Results}
\noindent\textbf{Comparison with Training-Free LLM Agents.} 
We perform a comprehensive comparison between our proposed method and existing training-free LLM agents, as summarized in Table~\ref{tab:main1}. 
To ensure a fair comparison, we adopt the same vision-language models across all methods: we use LaViLa~\citep{zhao2023learning} on EgoSchema to generate captions for 1-second video clips, CogAgent~\citep{hong2024cogagent} on NExT-QA, and LLaVA-NeXT~\citep{liu2024llavanext} on both Video-MME and MovieChat. 
The results demonstrate that our method consistently outperforms existing approaches in terms of both \textbf{efficiency} and \textbf{accuracy}.

Specifically, compared to global dense captioning methods like LLoVi \citep{zhang2023simple}, VideoAgent \citep{fan2024videoagent}, and DrVideo \citep{ma2024drvideo}, our method's advantages primarily lie in its efficiency. For example, on Video-MME, LLoVi and DrVideo require captions for at least 492 frames, while our method achieves better accuracy with an average of only 18.3 frames, significantly reducing the number of frames that need to be processed. This efficiency is attributed to our \textit{Perception-Verification} mechanism, which selectively identifies key frames, thus avoiding redundant global frame processing and ensuring comprehensive coverage of task-relevant information for accurate question answering. Additionally, our method also surpasses global processing approaches in terms of accuracy. On EgoSchema, our method achieves 6.4\% and 2.8\% higher accuracy than VideoAgent and DrVideo, respectively. This improvement is mainly due to our method's ability to eliminate redundant information and caption hallucinations, which otherwise interfere with LLM reasoning.

In comparison to the similarity-based key frame selection approach used by VideoAgent*~\citep{wang2024videoagent}, our method requires only 18.3 frames on the long video dataset Video-MME, demonstrating superior efficiency. This is because our method more accurately captures frames causally relevant to the task. While our method uses slightly more frames on shorter-duration datasets like EgoSchema and NExT-QA, it achieves 9.0\% and 5.5\% higher accuracy, respectively. This is primarily because our method extracts spatiotemporal information at multiple granularities and eliminates hallucinations in captions.
Compared to the clustering-based key frame selection method in VideoTree~\citep{wang2025videotree}, our method requires significantly fewer frames. This is because our method efficiently captures key frames without relying on global clustering and can better leverage frame-specific information rather than solely depending on captions.

To ensure a fair comparison, we conduct experiments using the same version of the open-source LLM, Mistral-8x7B~\citep{jiang2024identifying}, as shown in Table~\ref{tab:open_source_llms}.
The results demonstrate that our method outperforms the state-of-the-art method, DrVideo, in both accuracy and efficiency. 
This superiority stems from the proposed framework itself, rather than the inherent capability of the underlying LLM.

\noindent\textbf{Comparison with Proprietary and Open-Source MLLMs.} 
We compare our method with state-of-the-art proprietary and open-source MLLMs on EgoSchema and the long split of Video-MME, as shown in Table~\ref{tab:EgoSchema_all}. By incorporating the more powerful VLMs InternVL2.5-8B~\citep{chen2024expanding} and Qwen2.5-VL-7B~\citep{bai2025qwen2} as components, our method achieves accuracy comparable to Gemini~1.5-Pro and GPT-4o while using only a minimal number of frames, surpassing the standalone Qwen2.5-VL-7B and outperforming VideoChat-A1~\citep{wang2025videochat}, which also adopts the same VLMs—InternVL2.5-8B and Qwen2.5-VL-7B—as its components. This performance gain stems from the collaborative design of LLMs and MLLMs, which circumvents the inherent context-window limitations of a single MLLM and enables adaptive, multi-granularity information extraction.

\begin{table}[h]
\centering
\caption{Ablation results of Perception Tools on the NExT-QA subset. 
C, T, and D denote accuracy (\%) on the causal, temporal, and descriptive subsets, respectively.}
\label{tab:ablation_components}

\resizebox{\linewidth}{!}{
\begin{tabular}{cccccccc}
\toprule
Type & DS & TF & SF & C & T & D & Avg. \\
\midrule
1 & $\checkmark$ & $\times$      & $\times$      & 74.0 & 63.0 & 79.0 & 72.0 \\
2 & $\checkmark$ & $\checkmark$  & $\times$      & 77.5 & 66.0 & 79.5 & 74.3 \\
3 & $\times$     & $\checkmark$  & $\checkmark$  & 73.0 & 62.5 & 80.0 & 71.8 \\
\rowcolor{gray!20}
4 & $\checkmark$ & $\checkmark$  & $\checkmark$  & \textbf{79.0} & \textbf{67.0} & \textbf{84.5} & \textbf{76.8} \\
\midrule
5 & Uniform-k & $\checkmark$ & $\checkmark$ & 74.5 & 63.5 & 81.5 & 73.2 \\
6 & top-k     & $\checkmark$ & $\checkmark$ & 76.5 & 65.5 & 84.0 & 75.3 \\
\bottomrule
\end{tabular}
}
\end{table}

\subsection{Ablation Study}

\textbf{Ablation of the Perception Tools.} 
We perform ablation studies of the proposed Perception Tools on the NExT-QA subset, as shown in Table~\ref{tab:ablation_components}. The results indicate that \texttt{divergent search} (DS), \texttt{temporal focus} (TF), and \texttt{spatial focus} (SF) are complementary. Specifically, DS provides general reasoning cues and consistently benefits all question types. TF is particularly effective for temporal reasoning, such as identifying previous or subsequent events. SF facilitates the recognition of objects, attributes, and counts, which is especially valuable for descriptive questions.

We also compare strategies for \texttt{divergent search}, including uniform $k$-frame sampling, similarity-based top-$k$, and our watershed strategy (Types 4–6). Results show that the watershed strategy significantly improves causal and temporal tasks by capturing a broader range of relevant frames. 
We also provide a visualization analysis of the \texttt{divergent search} algorithm, and the details are presented in the Appendix.

\begin{table}[h]
\centering
\caption{Ablation results on the agentic framework on EgoSchema.}
\label{tab:ablation_agent}
\begin{tabular}{lcc}
\toprule
Framework & Acc. & Frames \\
\midrule
ReAct~\citep{yao2023react}        & 63.2 & 14.7 \\
Ours w/o Verification             & 64.8 & \textbf{8.1} \\
Ours                              & \textbf{69.2} & 11.2 \\
\bottomrule
\end{tabular}
\end{table}

\textbf{Ablation of Agentic Framework.} We evaluate the contribution of the proposed \textit{Perception-Verification} loop by fixing the action space and ablating the framework, as shown in Table~\ref{tab:ablation_agent}. We first highlight the impact of the verification mechanism: incorporating \textit{Key Observation Verification} yields a significant 4.4\% accuracy improvement, demonstrating that it effectively mitigates the misleading effects of hallucinated information, albeit with a slight overhead of additional frames. Furthermore, compared to the standard ReAct~\citep{yao2023react} baseline, integrating the \textit{Evaluation and Planning} module substantially reduces the average number of frames from 14.7 to 8.1, underscoring its crucial role in streamlining clue discovery and enhancing overall efficiency.

\begin{table}[h]
\centering
\caption{Comparison using an open-source LLM and ablation across open-source and proprietary LLMs on EgoSchema.}\label{tab:open_source_llms}
\begin{tabular}{lccc}
    \toprule
    Method & LLM & Acc. & Frames \\ 
    \midrule  
    DrVideo        & Mistral-8x7B & 47.6 & >90 \\
    \rowcolor{gray!20}
    Ours           & Mistral-8x7B & \textbf{51.8} & \textbf{14.9} \\
    \midrule 
    Ours           & GPT-3.5 & 66.2 & 13.3 \\ 
    Ours           & DeepSeek-V3 & 68.6 & 12.7 \\    
    \rowcolor{gray!20}
    Ours           & GPT-4 & \textbf{69.2} & \textbf{11.2} \\
    \bottomrule
\end{tabular}
\end{table}

\textbf{Ablation of LLMs.} We evaluate our framework with both open-source and proprietary LLMs (Table~\ref{tab:open_source_llms}). Results show that our prompting strategy transfers well across different LLMs. 
While absolute performance varies with the reasoning ability of each LLM, the key contribution lies in the design of the agentic framework.

\begin{figure}[h]
    \centering
    \includegraphics[width=0.6\linewidth]{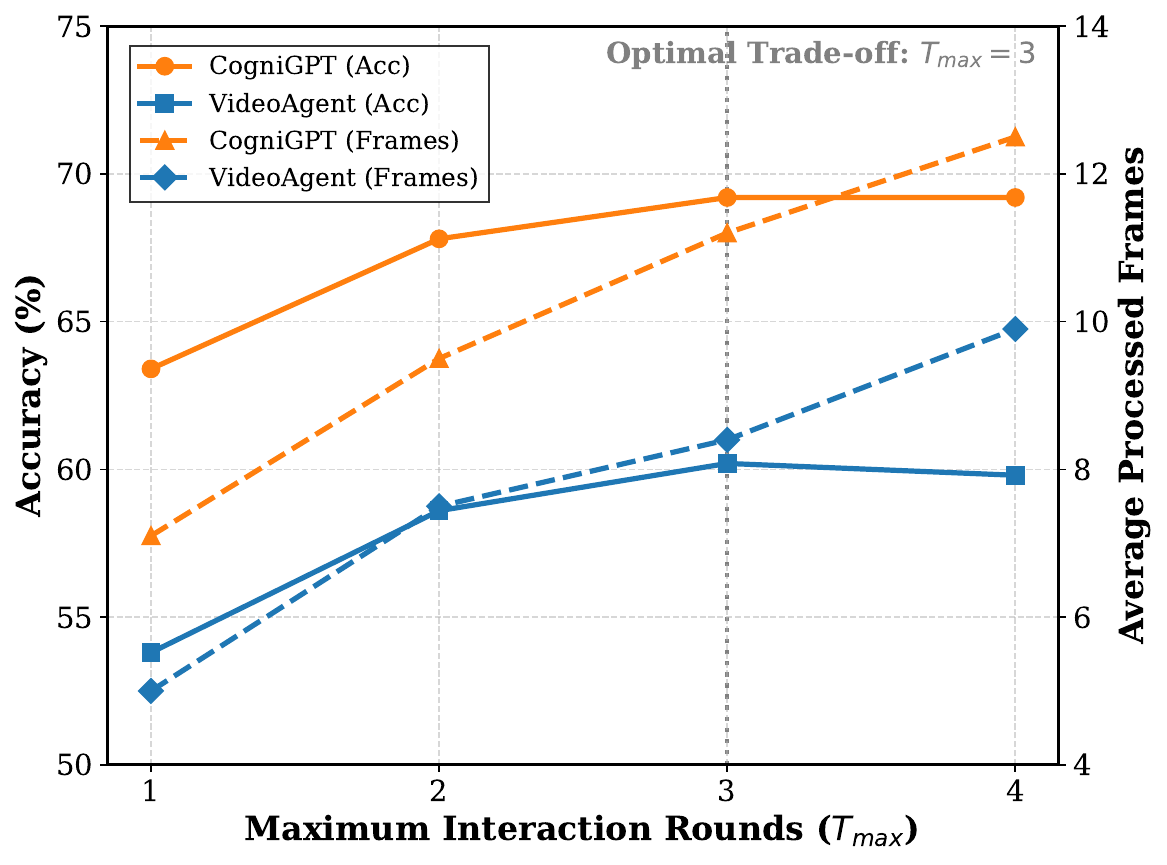}
    \caption{Ablation of interaction rounds ($T_{max}$).}
    \label{fig:ablation_tmax}
\end{figure}

\begin{figure*}[t]
    \centering
    \includegraphics[width=0.85\linewidth]{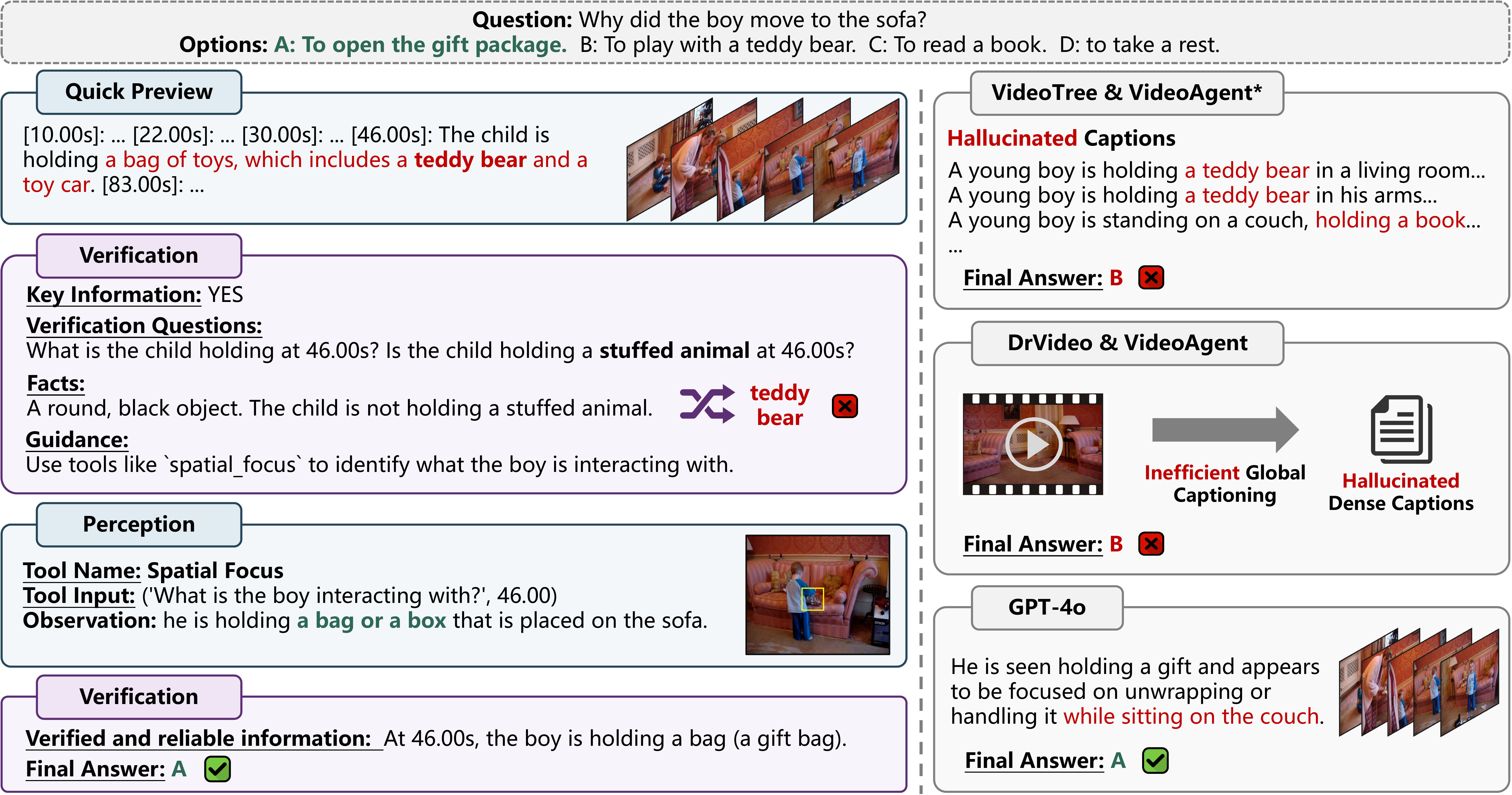}
    \caption{A case study from NExT-QA. CogniGPT progressively explores clues while effectively avoiding interference from hallucinated captions. In contrast, existing training-free LLM agents are misled by such hallucinations, and the global captioning methods of DrVideo and VideoAgent are inefficient. Although GPT-4o provides the correct answer, its reasoning process is affected by hallucinations.}
    \label{case}
\end{figure*}

\textbf{Ablation of Interaction Rounds.} We evaluate the impact of maximum interaction rounds ($T_{max}$) on accuracy and processed frames. As Figure~\ref{fig:ablation_tmax} shows, $T_{max}=3$ emerges as the optimal trade-off between accuracy and efficiency. Furthermore, empowered by our multi-granular perception and verification mechanisms, CogniGPT consistently outperforms VideoAgent across all iteration settings.

\begin{table}[h]
\centering
\caption{Experimental results on Haystack-LVBench. All retrieval metrics are reported in (\%).}
\label{tab:haystack_lvbench_results}
\resizebox{\linewidth}{!}{
\begin{tabular}{lcccc}
\toprule
Method & Frames & Precision & Recall & F1 \\
\midrule
VideoAgent*           & 10.1 & 1.2 & 8.5 & 2.1 \\
T*~\citep{ye2025re}   & 8.0  & 1.6 & 7.1 & 2.5 \\
Ours w/o Verification & \textbf{7.8} & 2.3 & 8.9 & 3.7 \\
Ours                  & 10.7 & \textbf{2.8} & \textbf{10.6} & \textbf{4.4} \\
\bottomrule
\end{tabular}
}
\end{table}

\textbf{Evaluation of Key-Information Grounding.} Furthermore, to verify that CogniGPT indeed selects key information and to assess the contribution of the verification mechanism to this process, we conduct a Needle-in-a-Haystack task on Haystack-LVBench~\citep{ye2025re}, as shown in Table~\ref{tab:haystack_lvbench_results}. We evaluate the retrieval performance of task-relevant key frames using Precision, Recall, and F1. The results show that, despite using fewer frames, our retrieval performance is superior, indicating that our method grounds key information more effectively than VideoAgent* and T*~\citep{ye2025re}. This advantage arises from our multi-granularity information selection strategy. In addition, the verification mechanism further strengthens key-information grounding by effectively preventing misleading caption hallucinations.

\begin{figure*}[h]
    \centering
    \includegraphics[width=0.85\linewidth]{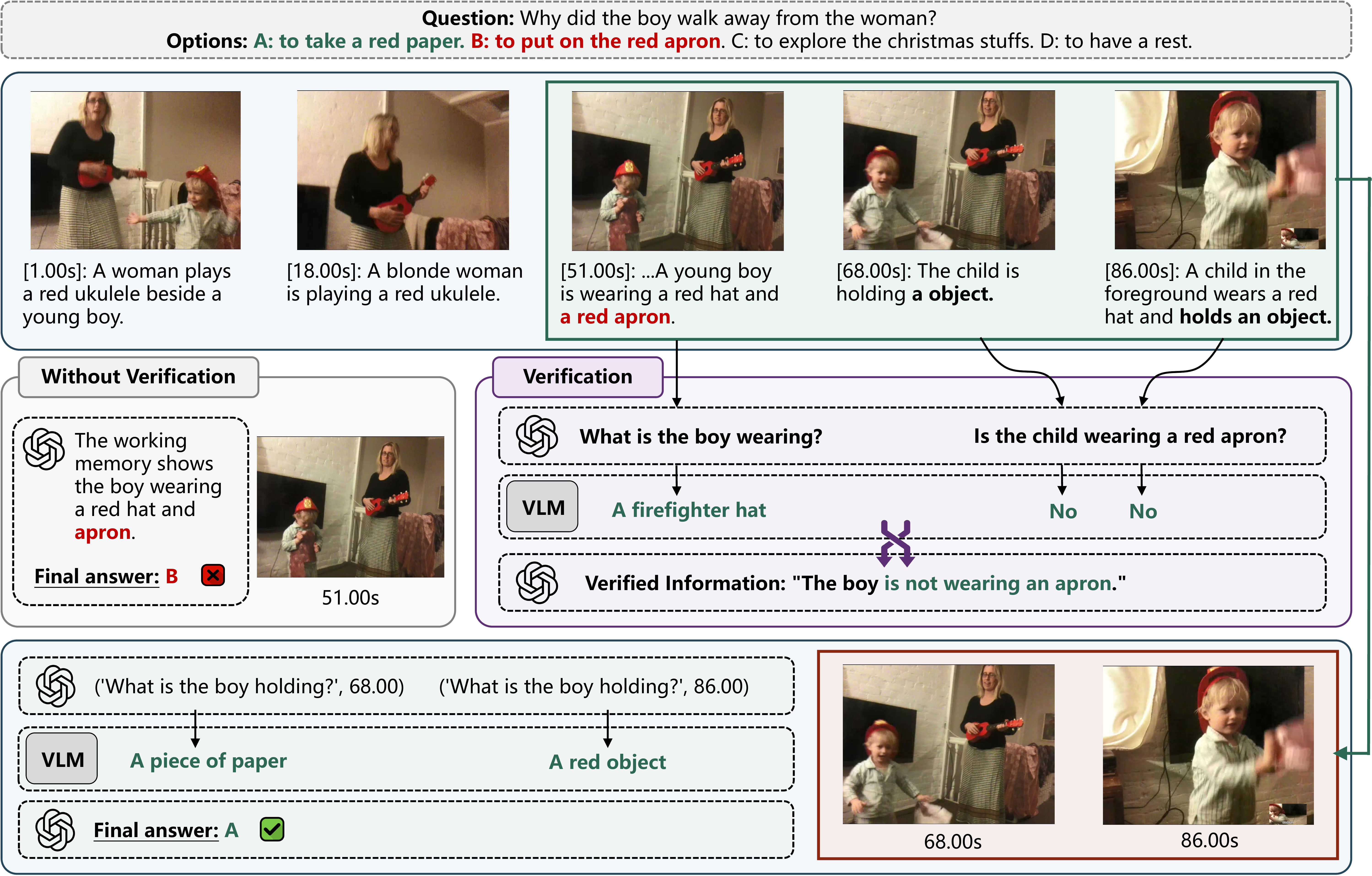}
    \caption{A case demonstrating the process of key information selection and verification. CogniGPT, through the collaboration between the LLM and VLM, eliminates the hallucination ``a red apron'' in the original caption by performing cross-validation across multiple frames and differently phrased questions. This mechanism further enables CogniGPT to localize the key frame and obtain the correct answer A. In contrast, without the verification mechanism, it incorrectly selects answer B.}
    \label{case_verification}
\end{figure*}

\subsection{Case Study}

\textbf{Robust Reasoning via Fine-Grained Perception.} Figure~\ref{case} illustrates a case from the NExT-QA dataset. In this example, CogniGPT first employs the Quick Preview to extract frames depicting the main scene. The Active Verification Agent then identifies the frame at 46.00s as a keyframe and detects that the object ``teddy bear'' is a hallucination. Subsequently, the Multi-Granular Perception Agent performs fine-grained spatial reasoning on the identified keyframe, determining that the true target object is a ``bag,'' and thus selects the correct answer (A). In contrast, existing training-free LLM agents are misled by hallucinated captions, leading them to select (B). Additionally, DrVideo and VideoAgent's dense captioning methods are inefficient. Although GPT-4o provides the correct answer, its reasoning process is affected by hallucinations. 

\textbf{Hallucination Elimination via Active Cross-Verification.} Figure~\ref{case_verification} presents a case illustrating the process of key information selection and verification. CogniGPT identifies and mitigates the task-related hallucination ``a red apron'' in the caption by cross-validating differently phrased questions across multiple candidate key frames, thereby preventing the hallucinated caption from misleading LLM reasoning. This mechanism further enables CogniGPT to accurately localize the true key information and produce reliable answers. We also provide failure cases in Appendix, including both the perception and verification stages.

\begin{table*}[t]
\centering
\caption{Average runtime, number of LLM calls, and token analysis per sample on EgoSchema.
Retrieval includes similarity computation and clustering. Time is in seconds.}
\label{tab:runtime_analysis}
\resizebox{0.9\textwidth}{!}{
\begin{tabular}{lccccccccccc}
\toprule
\multirow{2}{*}{Method}
& \multirow{2}{*}{\makecell{Embedding\\Time}}
& \multirow{2}{*}{\makecell{Retrieval\\Time}}
& \multirow{2}{*}{\makecell{Caption\\Time}}
& \multirow{2}{*}{\makecell{QA\\Time}}
& \multicolumn{3}{c}{LLM}
& \multirow{2}{*}{\makecell{Total\\Time}}
& \multirow{2}{*}{Frames}
& \multirow{2}{*}{Acc.} \\
\cmidrule(lr){6-8}
& & & & & Time & Calls & Tokens & & & \\
\midrule
VideoAgent* & 4.0 & 0.2 & 10.8 & -- & 84.9 & 10.0 & 1131.0 & 99.9 & \textbf{8.4} & 60.2 \\
VideoTree   & 4.0 & 0.2 & 80.5 & -- & 16.5 & 5.9 & 291.7 & 101.2 & 62.4 & 66.2 \\
DrVideo     & 0.6 & 0.1 & 118.5 & 2.4 & 21.7 & 5.0 & 292.0 & 143.3 & $>90$ & 66.4 \\
\rowcolor{gray!20}
Ours        & 4.0 & 0.1 & 10.3 & 1.5 & 23.6 & 6.8 & 392.9 & \textbf{39.5} & 11.2 & \textbf{69.2} \\
\bottomrule
\end{tabular}
}
\end{table*}

\subsection{Runtime Analysis}\label{sec: Runtime Analysis}

To assess the efficiency of our approach, we evaluate the per-sample average runtime, LLM calls, and generated tokens on the EgoSchema dataset using a single NVIDIA A6000 GPU (48GB). The runtime includes video/text embedding, similarity computation/clustering (retrieval), captioning, QA, and LLM responses. Compared to VideoTree and DrVideo, our method is significantly more efficient with fewer caption frames. When compared to VideoAgent*, we further reduce LLM inference time. This superior efficiency stems directly from our interactive perception-verification loop: MPA adaptively targets minimal task-relevant frames to drastically cut redundant captioning overhead, while AVA precisely cross-verifies these clues, bypassing the prolonged LLM reasoning cycles often triggered by rigid retrieval or hallucinations.

\section{Conclusion}

In this paper, we propose CogniGPT, an interactive framework that establishes a robust perception-verification loop for long video understanding. Breaking away from fixed, single-granularity pipelines, our MPA adaptively adjusts perception strategies and granularities based on task complexity. To ensure reasoning reliability, our AVA cross-verifies key observations across multiple perspectives to strictly eliminate vision-language hallucinations. Extensive experiments demonstrate that CogniGPT significantly outperforms existing training-free methods in both accuracy and efficiency.

\bibliographystyle{ACM-Reference-Format}
\bibliography{sample-base}

\newpage
\appendix

\begin{figure*}[t]
    \centering
    \includegraphics[width=0.66\textwidth]{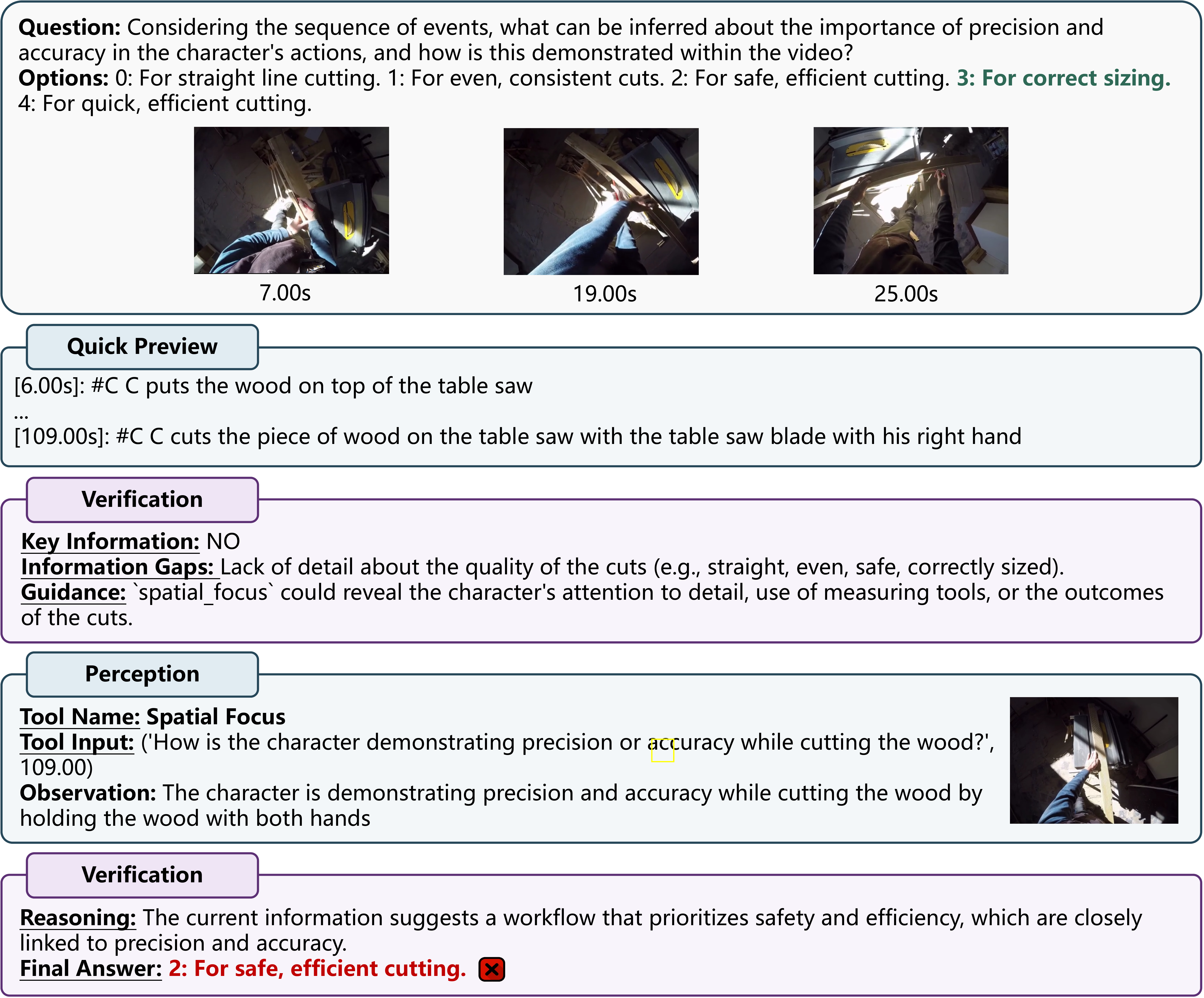}
    \caption{A failure case in the perception stage. The error arises because CogniGPT focuses solely on segments that are directly related to the question (e.g., the cutting process), while overlooking segments that are only potentially related to the question (e.g., the action of marking with a pen).}
    \label{failure_case}
\end{figure*}

\section{Prompt of LLM}

The LLM prompts for our Verification Agent, Perception Agent, and perception actions are as follows.

\definecolor{promptblue}{RGB}{52, 152, 219}
\definecolor{promptgreen}{RGB}{46, 204, 113}
\definecolor{promptorange}{RGB}{230, 126, 34}

\begin{tcolorbox}[
    breakable,
    colback=promptblue!10,
    colframe=promptblue,
    title={\textbf{Perception Agent Prompt}},
    fonttitle=\bfseries,
    boxrule=1pt,
    arc=3pt,
    left=5pt,
    right=5pt,
    top=5pt,
    bottom=5pt
]
\small
\texttt{You are a Perception Agent responsible for selecting the most appropriate perception tool to gather information based on the current understanding state and guidance.}

\vspace{3pt}
\texttt{Current question: \{question\}}\\
\texttt{Video duration: \{video\_duration\} seconds}

\vspace{3pt}
\texttt{Current working memory:}\\
\texttt{\{memory\}}

\vspace{3pt}
\texttt{Guidance for next action:}\\
\texttt{\{guidance\}}

\vspace{3pt}
\texttt{Available perception tools:}\\
\texttt{\{tools\}}

\vspace{3pt}
\textbf{IMPORTANT Notes:}
\begin{enumerate}
    \item \texttt{The segment captions with prefix `\#C' refer to the camera wearer, while those with prefix `\#O' refer to someone other than the camera wearer.}
    \item \texttt{Do NOT use single letters `C' or `O' as query for key\_frame\_selection tool. Instead, use specific objects, actions, or descriptive terms.}
\end{enumerate}

\vspace{3pt}
\texttt{You MUST output ONLY the following format with NO additional text or explanations:}\\
\texttt{Tool Name: [selected tool name]}\\
\texttt{Tool Input: [tool input parameters]}
\end{tcolorbox}

\begin{tcolorbox}[
    breakable,
    colback=promptgreen!10,
    colframe=promptgreen,
    title={\textbf{Verification Agent Step 1: Verification Prompt}},
    fonttitle=\bfseries,
    boxrule=1pt,
    arc=3pt,
    left=5pt,
    right=5pt,
    top=5pt,
    bottom=5pt
]
\small
\texttt{You are a Verification Agent analyzing the latest observation for task-relevant information.}

\vspace{3pt}
\texttt{Question: \{question\}}\\
\texttt{Video duration: \{video\_duration\} seconds}\\
\texttt{Latest observation: \{latest\_observation\}}

\vspace{3pt}
\texttt{Note: `\#C' refers to camera wearer, `\#O' refers to others.}

\vspace{3pt}
\texttt{Analyze the observation:}
\begin{enumerate}
    \item \texttt{Does it contain information that is critical for answering the question?}
    \item \texttt{If yes, what key information requires verification? Generate two to three distinct verification questions with timestamps, to assess whether the key information is hallucinated.}
\end{enumerate}

\vspace{3pt}
\textbf{Important Guidelines for Verification Questions:}
\begin{itemize}
    \item \texttt{Keep questions SIMPLE and DIRECT - avoid complex or compound questions}
    \item \texttt{Focus on basic visual facts that can be easily verified (colors, objects, actions, positions)}
\end{itemize}

\vspace{3pt}
\texttt{Verification Questions Examples:}\\
\texttt{Verification Questions: [(``Is the boy's shirt red?'', 15.2), (``What color is the boy's shirt?'', 15.2)]}

\vspace{3pt}
\texttt{CRITICAL: You MUST output ONLY the following format with NO additional text or explanations:}\\
\texttt{Key Information: [YES/NO]}

\vspace{3pt}
\texttt{If YES:}\\
\texttt{Verification Questions: [(``question1'', timestamp1), (``question2'', timestamp2)]}
\end{tcolorbox}

\begin{tcolorbox}[
    breakable,
    colback=promptorange!10,
    colframe=promptorange,
    title={\textbf{Verification Agent Step 2: Sufficiency Assessment Prompt}},
    fonttitle=\bfseries,
    boxrule=1pt,
    arc=3pt,
    left=5pt,
    right=5pt,
    top=5pt,
    bottom=5pt
]
\small
\texttt{You are a Verification Agent. Analyze working memory to determine information sufficiency.}

\vspace{3pt}
\texttt{Current question: \{question\}}\\
\texttt{Video duration: \{video\_duration\} seconds}

\vspace{3pt}
\texttt{Current working memory:}\\
\texttt{\{working\_memory\}}

\vspace{3pt}
\texttt{Notes:}
\begin{enumerate}
    \item \texttt{`\#C' = camera wearer, `\#O' = other person}
    \item \texttt{All information is verified/reliable}
\end{enumerate}

\vspace{3pt}
\textbf{Information Sufficiency Assessment}
\begin{enumerate}
    \item \texttt{Information sufficiency: Is collected information sufficient to answer reliably?}
    \item \texttt{Information gaps: What key information is missing?}
    \item \texttt{Next step decision: Continue gathering or terminate with answer?}
\end{enumerate}

\vspace{3pt}
\texttt{Criteria:}
\begin{itemize}
    \item \texttt{CONTINUE: Critical info missing, insufficient, low confidence}
    \item \texttt{TERMINATE: Sufficient relevant info gathered}
\end{itemize}

\vspace{3pt}
\texttt{Output (be concise):}\\
\texttt{Analysis: [brief assessment of sufficiency and gaps]}\\
\texttt{Decision: [continue/terminate]}

\vspace{3pt}
\texttt{If CONTINUE:}\\
\texttt{Guidance: [what to gather next, which tools]}

\vspace{3pt}
\texttt{If TERMINATE:}\\
\texttt{Final Answer: [number 0-4]}
\end{tcolorbox}

\begin{tcolorbox}[
    breakable,
    colback=purple!5!white,
    colframe=purple!75!black,
    title=Multi-Granular Perception Actions,
    fonttitle=\bfseries,
    boxrule=1pt,
    arc=3pt,
    left=5pt,
    right=5pt,
    top=5pt,
    bottom=5pt
]
\small
\textbf{1. divergent\_search}

Find video segments related to a query within a broad time range and generate rough descriptions.

\textbf{TOOL INPUT FORMAT:} \texttt{('query\_text', (start\_time, end\_time))}

\textbf{Input must be:} \texttt{('man with glasses', (150.0, 315.0))}

\textbf{EXAMPLE:} \texttt{('person', (0.0, 90.0))}

Returns top-k most relevant segments with timestamps and rough descriptions.

\vspace{0.5cm}

\textbf{2. spatial\_focus}

Analyze spatial relationships and visual attributes at specific time points in the video.

\textbf{TOOL INPUT FORMAT:} \texttt{[('question\_text1', time\_point1), ('question\_text2', time\_point2), ...]}

\textbf{EXAMPLE:} \texttt{[('What objects are visible in the scene?', 10.5), ('What color is the car?', 20.3)]}

\textbf{NOTE:} Specializes in understanding scene composition, object attributes, and spatial relationships.

\vspace{0.5cm}

\textbf{3. temporal\_focus}

Identify key scenes within specific time intervals and generate captions.

\textbf{TOOL INPUT FORMAT:} \texttt{[(start\_time1, end\_time1), (start\_time2, end\_time2), ...]}

\textbf{EXAMPLE:} \texttt{[(10.0, 30.0), (37.0, 47.5), (70.0, 78.0)]}

Returns timestamps and captions of the most representative scenes in the given time ranges.

\end{tcolorbox}

\section{Failure Case Analysis}\label{sec: Failure Case Analysis}

\begin{figure*}[t]
    \centering
    \includegraphics[width=0.66\textwidth]{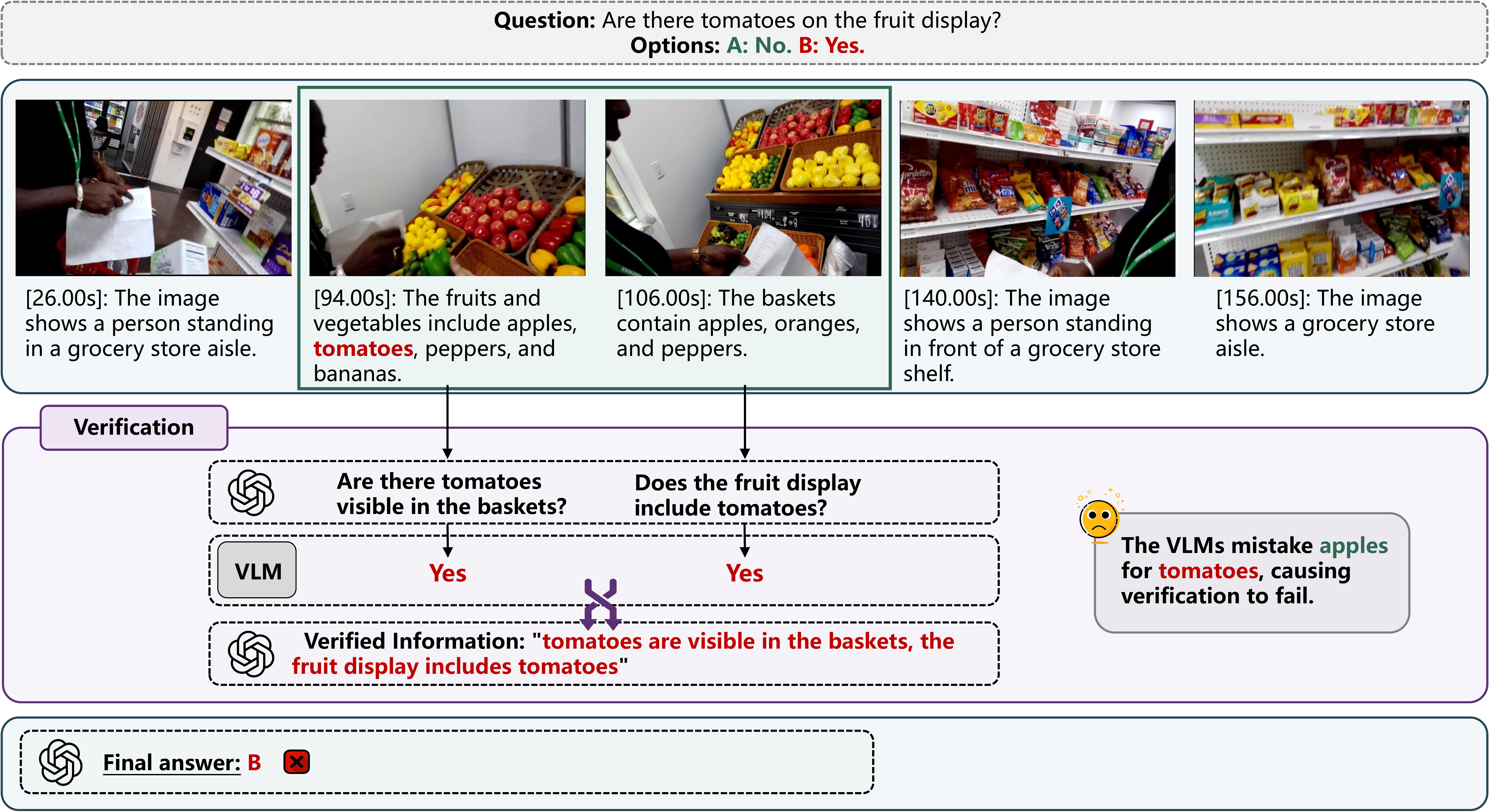}
    \caption{A failure case in the verification stage. The error arises because the VLM is limited by its recognition capability. In extreme cases, the VLM may hallucinate a consistent answer across multiple frames and the question (e.g., repeatedly judging that there are tomatoes in several frames), which leads to verification failure.}
    \label{failure_case_verification}
\end{figure*}

We present failure cases in both the \textbf{perception stage} and the \textbf{verification stage} from the EgoSchema dataset, as shown in Figure~\ref{failure_case} and Figure~\ref{failure_case_verification}.

First, we present a challenging case involving long-range implicit causal reasoning during the \textbf{perception stage}, as shown in Figure 1. Answering the question requires associating the explicitly queried action (the cutting process at 109.00s) with a semantically distinct prerequisite action (the pen-marking process between 6.00s and 25.00s). To achieve high efficiency, CogniGPT's Divergent Search mechanism is deliberately designed to filter out segments with low semantic relevance, extracting only a minimal, highly reliable set of clues. Consequently, when a causal premise and the queried action exhibit a massive semantic and temporal gap, the prerequisite segment may fall below our adaptive perception threshold. Ultimately, rather than a design flaw, this case highlights a fundamental trade-off in training-free agents: balancing optimal operational efficiency against the computational cost of exhaustive, open-ended causal exploration. Addressing this balance remains a key direction for future research.

Secondly, we present a failure case of the \textbf{verification stage}, as shown in Figure~\ref{failure_case_verification}. Specifically, for the hallucinated term “tomatoes” in the caption at 94.00s, although multiple frames and questions are used for cross-verification, the VLM’s limited recognition capability causes consistent errors across several frames and queries under certain extreme conditions (in this example, the model repeatedly misidentifies an apple as tomatoes across multiple frames). Consequently, the hallucination is not successfully corrected. Addressing this issue requires improving the perceptual capacity of MLLMs themselves to avoid such consistency-induced hallucinations, which warrants further investigation in future work.

\begin{figure}[h]
    \centering
    \begin{minipage}{0.42\linewidth}
        \centering
        \includegraphics[width=\linewidth]{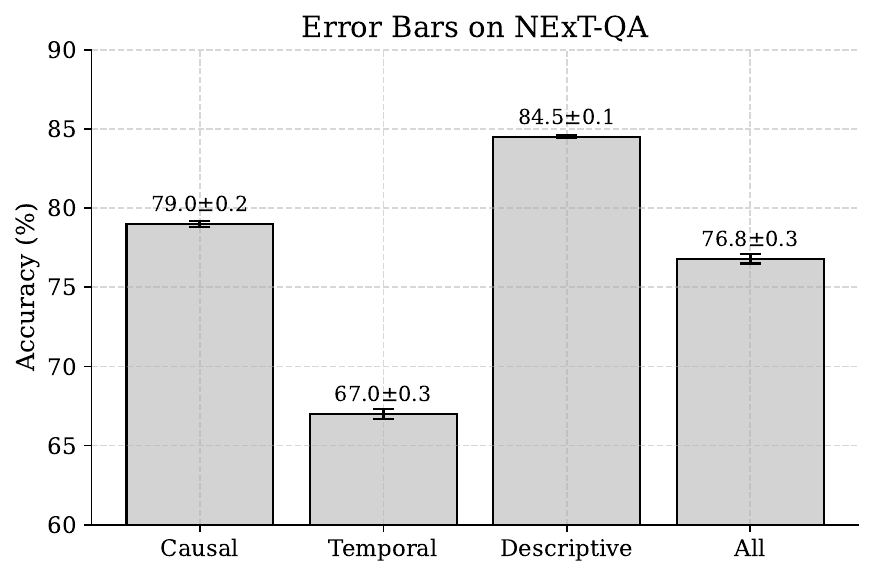}
    \end{minipage}\hspace{0.04\linewidth}
    \begin{minipage}{0.42\linewidth}
        \centering
        \includegraphics[width=\linewidth]{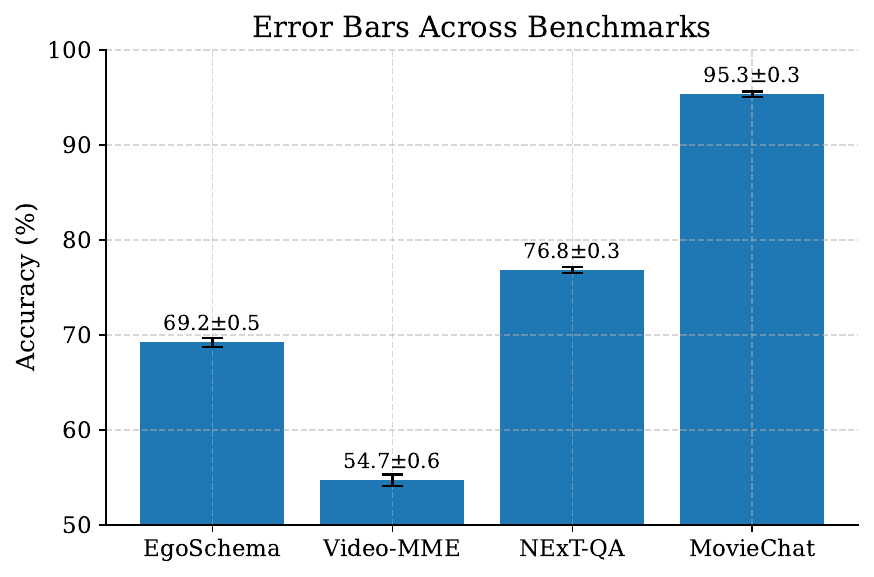}
    \end{minipage}

    \caption{Error bar analysis across benchmarks.
    Left: Error bar analysis on the NExT-QA benchmark. C, T, D, and All denote the accuracy (\%) on the causal, temporal, and descriptive subsets, and their overall average, respectively.
    Right: Error bars across EgoSchema, Video-MME, NExT-QA, and MovieChat benchmarks.}
    \label{error_analysis}
\end{figure}

\section{Error Analysis}

We conduct an error bar analysis on the subsets of the NExT-QA benchmark. Specifically, we run our model 10 times under identical experimental conditions and calculate the mean and standard deviation of accuracy. As shown on the left side of Figure~\ref{error_analysis}, the standard deviations for the causal, temporal, and descriptive questions are 0.2\%, 0.3\%, and 0.1\%, respectively. These results demonstrate that our method exhibits strong robustness and stability across different question types.

Additionally, we perform five experiments with different random seeds on the EgoSchema, Video-MME, NExT-QA, and MovieChat benchmarks. The error bars are shown on the right side of Figure~\ref{error_analysis}. The standard deviations on the four datasets are 0.5\%, 0.6\%, 0.3\%, and 0.3\%, respectively. These findings highlight the robustness and generalization ability of our method across datasets with diverse contexts and temporal scales, ensuring the statistical significance and reliability of the reported results.

\section{Visualization Analysis of Divergent Search}\label{sec:divergent_visualization}

\begin{figure}[h]
    \centering
    \includegraphics[width=0.3\textwidth]{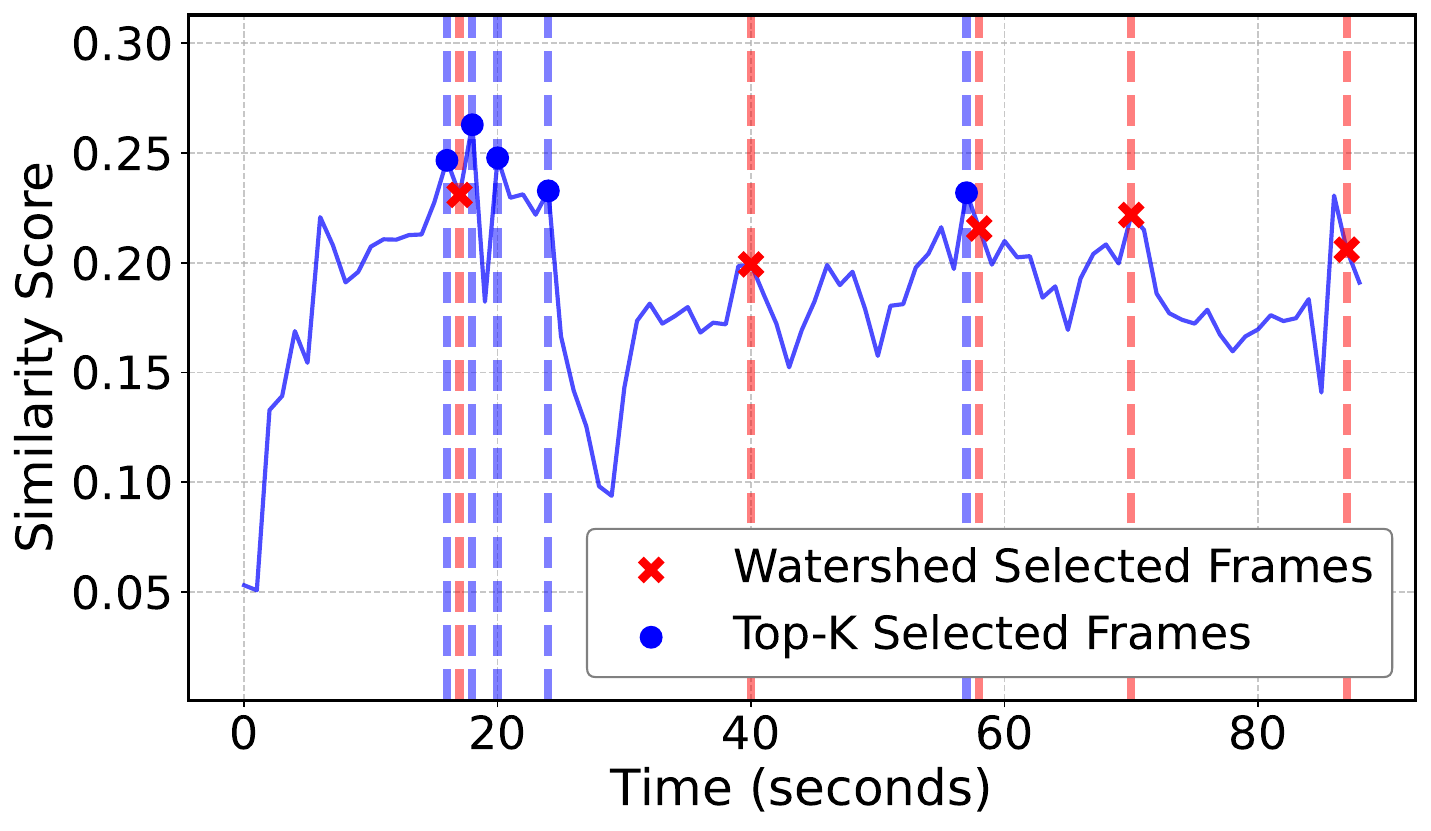}
    \caption{Comparison of divergent search strategies on NExT-QA.}
    \label{fig:similarity_plot}
\end{figure}

We conduct a visualization analysis of the \texttt{divergent search} algorithm. As shown in Figure~\ref{fig:similarity_plot}, unlike the commonly used top-$k$ strategy in VideoAgent~\citep{fan2024videoagent}, which often selects highly similar and redundant frames, our watershed strategy produces more diverse and informative frame selections.

\section{Dataset Details}

We conduct experiments on four video understanding benchmarks: EgoSchema~\citep{mangalam2023egoschema}, Video-MME~\citep{fu2024video}, NExT-QA~\citep{xiao2021next}, and MovieChat~\citep{song2024moviechat}.

\textbf{EgoSchema} contains first-person perspective videos (avg. 3 min) of daily activities, with multiple-choice questions requiring temporal understanding. We use the official subset of 500 labeled samples for fair comparison.

\textbf{Video-MME} is a recently introduced benchmark for evaluating MLLMs on comprehensive video understanding. In our experiments, we focus on its long-form subset, which consists of 300 open-domain videos (30–60 min, avg. ~41 min) annotated with 900 expert-crafted multiple-choice questions. This subset emphasizes long-range temporal reasoning over extended video contexts.

\textbf{NExT-QA} assesses temporal reasoning, causal analysis, and descriptive understanding, with videos averaging 44 seconds. We sample 600 instances (200 from each of the three question categories) for evaluation.

\textbf{MovieChat} consists of 1,000 movie/TV clips (average length: 10 minutes). We use the official test split of 170 samples and evaluate using the \textit{Global Mode}, which performs holistic analysis.

The datasets used in this experiment cover a wide range of video durations, question types, and reasoning tasks, ensuring a comprehensive evaluation of CogniGPT.

\end{document}